\documentclass[12pt, a4paper]{article}
\usepackage[utf8]{inputenc}
\usepackage[T1]{fontenc}
\usepackage{amsmath}
\usepackage{graphicx}
\usepackage[colorlinks=true, allcolors=blue]{hyperref}
\usepackage[margin=1in]{geometry}
\usepackage{natbib}  
\setcitestyle{authoryear,round}
\usepackage{booktabs}  
\usepackage{multirow}  
\usepackage{array}     
\usepackage{tabularx}
\usepackage{ragged2e}
\usepackage{graphicx}
\usepackage{doi}
\usepackage{url}
\usepackage{authblk} 
\usepackage{datetime} 
\usepackage{setspace} 

\title{\Large\textbf{AI, Global Governance, and Digital Sovereignty}}

\author[1]{Swati Srivastava\thanks{Corresponding author: srivas70@purdue.edu}}
\author[2,3,4]{Justin Bullock}

\affil[1]{\textit{Political Science, Purdue University}}
\affil[2]{\textit{Bush School of Government \& Public Service, Texas A\&M University}}
\affil[3]{\textit{Convergence Analysis}}
\affil[4]{\textit{Global Governance Institute}}

\date{\textit{Draft: \today}}

\begin{document}
\pagenumbering{gobble}
\maketitle

\begin{abstract}
This essay examines how Artificial Intelligence (AI) systems are becoming more integral to international affairs by affecting how global governors exert power and pursue digital sovereignty. We first introduce a taxonomy of multifaceted AI payoffs for governments and corporations related to instrumental, structural, and discursive power in the domains of violence, markets, and rights. We next leverage different institutional and practice perspectives on sovereignty to assess how digital sovereignty is variously implicated in AI-empowered global governance. States both seek sovereign control \textit{over} AI infrastructures in the institutional approach, while establishing sovereign competence \textit{through} AI infrastructures in the practice approach. Overall, we present the digital sovereignty stakes of AI as related to entanglements of public and private power. Rather than foreseeing technology companies as replacing states, we argue that AI systems will embed in global governance to create dueling dynamics of public/private cooperation and contestation. We conclude with sketching future directions for IR research on AI and global governance.
\end{abstract}

\vspace{1em}

\newpage
\pagenumbering{arabic}
\setcounter{page}{1}

\section{Introduction}

This essay examines how Artificial Intelligence (AI) systems are becoming more integral to international affairs by affecting how global governors exert power and invoke digital sovereignty. AI implicates governing power in different ways. For states, AI is an “enabling force” \citep[p.39]{horowitz2018} \citep[p. 960]{arsenault2024} with promises to improve capabilities and competitiveness \citep{bullock2024}. In 2022, U.S. federal agencies “reported about 1,200 AI use cases—specific challenges or opportunities that AI may solve” \citep{gao2023}. In light of this perceived promise, states are racing to secure AI as a national asset before their rivals \citep[p.924]{horowitz2024} and seeking to set the agenda in AI governance debates \citep{bradford2023, canfil2024}. Yet, states must also rely on globalized technology infrastructures of data, computing resources, and human talent that make zero-sum competition difficult \citep[p. 882]{ding2024}. Meanwhile, AI is core to the continued dominance of traditional Big Tech firms Alphabet, Meta, Amazon, and Microsoft \citep{zuboff2019, srivastava2023}. Newer players OpenAI and Anthropic have their foundation models integrated into everyday chatbots and AI assistants, while chip designer Nvidia and manufacturer TSMC have seen their fortunes soar. The centrality of private actors to AI advances have led to claims of an emerging “technopolar” order where “technology companies wield the kind of power in the domains once reserved by nation-states” \citep[p. 28]{bremmer2023}. In response, states have pursued schemes for “digital sovereignty” \citep{bellanova2022, broeders2023, adler2024}. 

But what is AI? Over the past 75 years, “artificial intelligence” has been used to describe various machine architectures. Work began on AI shortly after the invention of mechanical computers during the 1940s \citep{dyson2012}. Early approaches to AI included architectures of symbolic manipulation and statistical inference. Throughout the second half of the 20th century and the first decade of the 21st century, interest in and progress of AI systems waxed and waned with a series of “AI summers” followed by “AI winters.” During the same period, digitalization marched forward with the rise of personal computers and the world wide web \citep{dyson2012}. These developments highlighted the governance challenges and opportunities from vast increases in computation and the global interconnectedness of computation \citep{denardis2020}. Digital governance, for example, has become its own area of inquiry \citep{milakovich2012, luna2017, manoharan2023}. However, modern machine learning and neural network forms of AI became more capable in the early 2010s. The creation of “frontier” large language models over the last decade has ushered in AI systems with enhanced capabilities in a wide array of areas, including reading comprehension, image and speech recognition, and predictive reasoning \citep{kiela2023}. Human judgment in governance is by some accounts being both augmented and replaced by machine intelligence \citep{bullock2019, young2019}. At the same time, the ethical AI community warns about perpetuating “AI hype” centered on future risks \citep{bender2021} while ignoring current AI harms impacting marginalized groups \citep{eubanks2018, noble2018, benjamin2019}. Ultimately, as a “general purpose technology,” frontier AI is unlike a specific tool or weapon and more like electricity in its transformative potential \citep{horowitz2018}. 

What does the proliferation of frontier AI systems mean for global governance? International Relations (IR) has examined how control of data and computing resources entrenches state power \citep{farrell2023}, grows private power \citep{atal2021, srivastava2023}, and raises new dilemmas for human rights \citep{wong2023}, regulatory regimes \citep{gorwa2024}, and ethics \citep{erskine2024}. State use of AI tools for digital repression, such as censorship and surveillance, is perhaps most evident in China’s treatment of the Uyghurs, but the same technologies are exported globally and are also prevalent in democracies \citep{deibert2020, feldstein2021}. AI riches accrue to large corporations who are able to invest billions in computing power, data, and human talent \citep{vipra2023}, all the while operating without public accountability \citep{lehdonvirta2022}. As AI advancements create market concentrations, the regulatory gap widens \citep{culpepper2020, seidl2022}. 

Our aim in this agenda-setting essay is (1) to map how frontier AI systems equip public and private global governors with new ways of exercising power and (2) to assess the resulting implications for the emergence of digital sovereignty. We argue that AI systems do not affect power or sovereignty in singular ways. Instead, we create a taxonomy of AI payoffs related to instrumental, structural, and discursive power \citep{lukes1974, fuchs2013}. The purpose of the taxonomy is to take stock of the variety of ways in which AI systems have been, or might be, used by public and private governors (on varieties of global governors, see \citep{avant2010}. For public global governors such as states, we explore the development of powerful weapons with decreased human oversight (instrumental power), increases in internal control by supercharging surveillance capacity (structural power), and the ability to tailor propaganda to individual susceptibilities (discursive power). For private global governors such as corporations, we explore the invasive control of employees (instrumental power), concentration of computational resources (structural power), and alteration of the trust landscape (discursive power). We also discuss how autonomous AI models may express more agentic control over governance decision-making, including being in conflict with the goals of humans and organizations that create the AI systems. By including private governors and potential AI agents, our taxonomy moves beyond recent overviews of AI for international politics that are largely state-centric \citep[p.929-930]{horowitz2024} or skeptical of AI’s unforeseen transformative potential \citep[p.960]{arsenault2024}.

We present the digital sovereignty stakes of AI as related to entanglements of public and private power. Policy experts lament that “big technology firms have effectively become independent, sovereign actors in the digital realms they have created” \citep[p.28]{bremmer2023}, echoing claims by some IR scholars that online platforms now exhibit “virtual sovereignty” \citep{kelton2022}. Within this context, some states are pursuing “sovereign AI” in their national strategies, as seen in India’s assertion: “We are determined that we must have our own sovereign AI” \citep{barik2023}. But India’s ability to meet this objective is questionable \citep{panday2024}. Indeed, American firm Nvidia regards its chips as integral to state pursuit of sovereign AI, which the company defines as “a nation’s capabilities to produce artificial intelligence using \textit{its own} infrastructure, data, workforce and business networks” (\citep{nvidia2024} emphasis added). The French government also explicitly connects AI and sovereignty, recently acknowledging “our lag in the field of artificial intelligence undermines our sovereignty. Weak control of technology effectively implies a one-way dependence on other countries. In the privatized and ever-evolving field of AI, public power appears largely outmatched, limiting our collective ability to make choices aligned with our values and interests” \citep[p.8]{aicommission2024}. These efforts follow a broader resurgence of sovereignty talk in global discourse \citep{paris2020}. In reality, digital sovereignty has multifaceted meanings that authorize a range of policy practices \citep{bellanova2022, broeders2023, adler2024}. 

We build on this later work to discuss AI’s implications for digital sovereignty in two ways. As an international institution, sovereignty is state-centric \citep{onuf1991} and relies on states keeping nonstate actors out of their exclusive club \citep{barkin2021}. In the institutional perspective, the AI race creates opportunities for states to assert digital sovereignty over AI infrastructures and to be seen as taking on Big Tech as potential rivals. Recent European regulations such as the Digital Services Act, Digital Markets Act, and AI Act project Europe as an autonomous actor over private, particularly non-European, AI infrastructures. However, sovereignty is also an ongoing social practice \citep{wendt1992, biersteker1996}, where performing sovereign functions may require states working with nonstate actors \citep{srivastava2022a}. In the practice perspective, digital sovereignty is achieved through largely private AI infrastructures and depends on governments and companies co-developing capacity for AI innovation and regulation. Sticking with the European context, the Data Act and the Data Governance Act aim to monetize public data for European companies, while the Digital Services Act adopts a “co-regulatory” model with tech platforms. Thus, AI’s diverse power payoffs in global governance are likely to result in dual dynamics of states pursuing sovereignty over AI and sovereignty through AI.

The rest of the article proceeds as follows. The next section introduces the classic “three faces of power” approach in global governance. The third section maps how public and private governors may use AI to alter their instrumental, structural, and discursive power in the domains of violence, markets, and rights. We also consider how autonomous AI agents may be seen as decision-makers in global affairs. In the fourth section, we discuss AI’s implications for digital sovereignty from both the institutional and practice perspectives. The fifth section concludes with directions for future IR research.

\section{Three Faces of Power in Global Governance}
Global governance scholars have identified varieties of power \citep{barnett2005} beyond the standard reference to power as “A compels B to do something B would not otherwise do” \citep{dahl1957}. One influential source is Steven Lukes’ three faces of power \citeyear{lukes1974, lukes2005}. Lukes discusses compulsory influence as the first face of decision-making power, then brings in agenda-setting as the second face of non-decision-making power \citep{bachrach1962}, and identifies the ability to shape interests as the third face of ideological power. For Lukes, the second and third faces are less visible than the first. In a policy context, one might see the first face of power when veto players decide which proposals win or lose, but be less attuned to the second face in agenda-setting that influences which proposals even come for a vote or the third face in ideological forces that shape interests and desires for certain policies.

Lukes’ three faces of power have been operationalized in the global governance literature as instrumental, structural, and discursive power \citep{fuchs2013}. Instrumental power “analyzes direct, observable relationships of power deriving from actor-specific material resources, while structural material and discursive approaches situate power and its use in material and ideational institutions and structures” \citep[p.80]{fuchs2013}. Consider corporations, who exercise instrumental power through direct, observable interactions that result in influence over political decision-makers, for instance by lobbying through campaign finance contributions to influence politicians or public relations media strategies to influence consumers. Corporations also wield structural power through occupying structurally advantageous positions that grant them influence, for instance when their market share dominance affords agenda-setting privileges or to be seen as “too big to fail.” Finally, corporations engage in discursive power through creating and upholding dominant discourses that influence what others want (preferences), how others think of themselves (identity), how others see the corporations (reputation), and the cultural scripts undergirding society-corporate relations writ large. Examples include expecting fast shipping everywhere based on certain e-commerce deals or the burnout/self-care culture that only seeks to enhance worker productivity. 

Instrumental, structural, and discursive power all work together as mechanisms underlying the three faces of power. As such, they are complementary. For instance, consider Apple’s new Vision Pro virtual-reality smart glasses. To ensure that Vision Pro is successful, Apple could use lobbying to avoid regulatory scrutiny related to privacy (instrumental power), leverage its market dominance in consumer technologies to create seamless connections with iPhone, iWatch, and other Apple products (structural power), and activate its consumers’ desires to want Vision Pro for convenience and signaling purposes (discursive power). Although actors mobilize instrumental, structural, and discursive power simultaneously, it is still useful to understand their unique logics when mapping governance outcomes.

The three faces of power help clarify hidden risks for how power operates. Instrumental power may be the most observable and hence most easily operationalized into assessments of global governing power. Structural and discursive power are less observable. As such, AI-related global governance assessments benefit from taking a wider spectrum of power resources in instrumental, structural, and discursive power. Moreover, the three faces of power also invoke different repertoires. Instrumental and structural power might mobilize more material resources that could lead themselves to better oversight compared to the ideational resources of discursive power. For instance, we may be able to track corporate spending or anticompetitive practices in the first two faces of power, but it may be difficult to meaningfully capture corporate influence on identity and preferences in the third face of power. 

Finally, the three faces of power operate in distinct governance domains. Borrowing from how IR broadly organizes its subfields, we focus on domains of violence, markets, and rights \citep{srivastava2022b, adler2024}. We next apply the three faces of power in the domains to generate a taxonomy of how AI may be used by global governors.

\section{Taxonomy of AI Uses by Global Governors}
Global governors may use AI as sources of instrumental, structural, and discursive power in the domains of violence, markets, and rights. We catalogue some illustrative examples in this section, summarized in Table 1, to highlight the variety of power payoffs of AI in global governance.

\begin{figure}[h!]
\centering
\includegraphics[width=\textwidth]{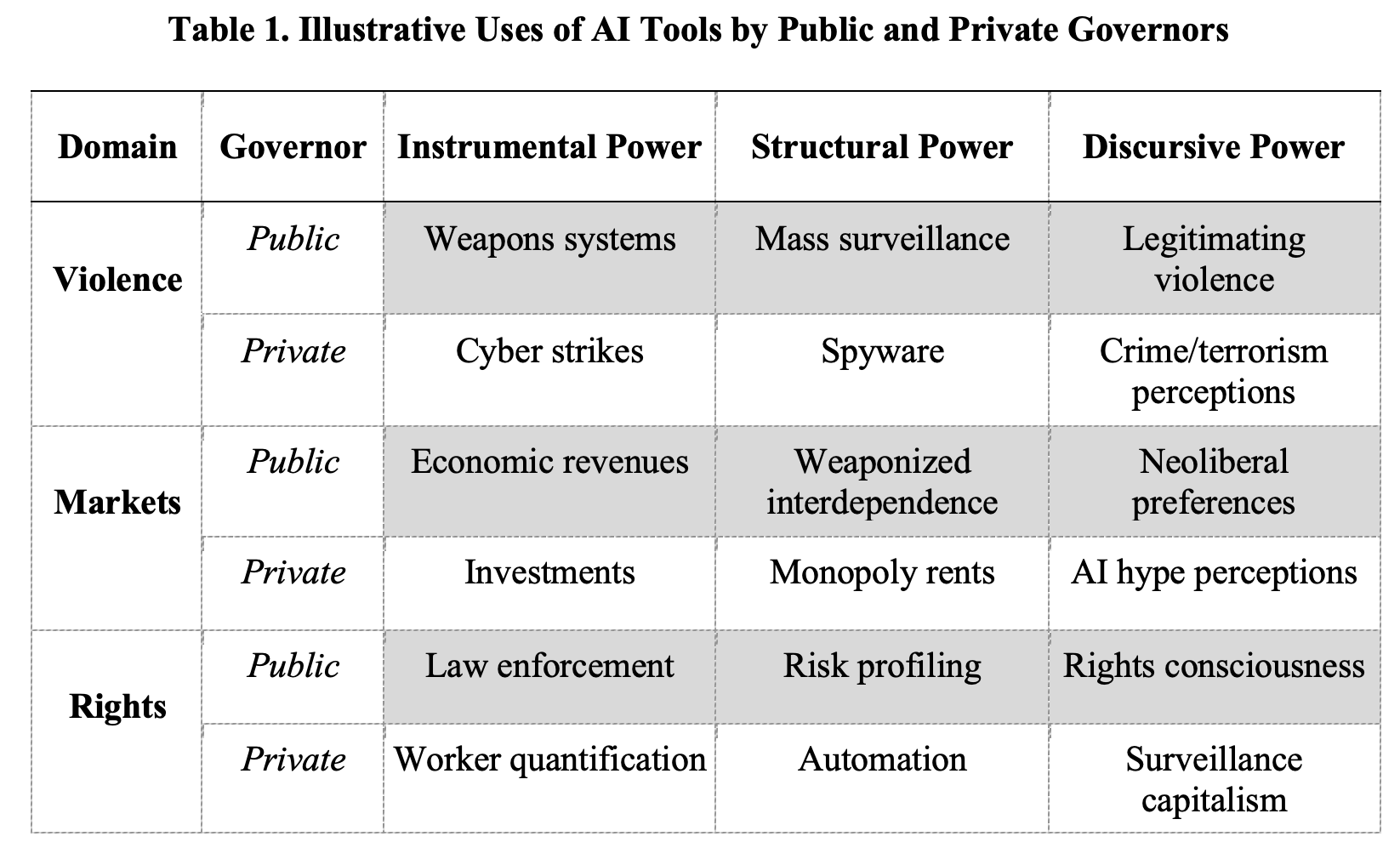} 
\end{figure}

\subsection{Public Authorities and Governance by AI}
Governments have already used AI technologies to create autonomous weapons to enhance their instrumental power. AI can autonomously and by augmentation of human capabilities increase the precision of weapons to harm targets and conduct attacks at a larger scale and from a further distance, all while reducing the likelihood of danger from one’s own forces by removing the need for them to be near the target \citep{canfil2024}. Autonomous weapons are a continuation of earlier digital warfare tools. AI can also increase the ability of both offensive and defensive tools in cyberspace to subvert the capabilities of an adversary. Structural power may be reflected in AI empowering states' capacity for surveillance of their populations, making subjects more legible and hence governable (Scott 1998). In this way, states’ structural position related to the governed is boosted by AI tools. Additionally, discursive power may be mobilized by the use of AI to shape citizen preferences to legitimate violence. For instance, governments may use AI models to generate highly personalized propaganda to distribute in information warfare campaigns \citep[p.918]{horowitz2024}. Similar to the use of radio and cable news by political operatives in major atrocities in Nazi Germany and Rwanda, generative AI can be deployed on a larger, more tailored scale to justify state-sanctioned violence. 

Governments wield significant authority in setting and enforcing the rules by which market actors are expected to compete. In instrumental power, states can use AI tools to increase economic revenues through enhancing tax enforcement capabilities or identifying fraudulent behavior in welfare provision, as is already being done with Medicare claims in the US \citep{bullock2020}. Governments may further privilege AI tools in financial governance infrastructures, for instance in determining interest rates or setting government loan terms, and use AI systems for government procurement of commercial services \citep{ahmadi2023}. In terms of structural power, states such as the U.S. can entrench their market dominance in AI against China through weaponizing economic interdependence by activating export controls in semiconductors and disincentivizing international academic collaboration through government grant allocation \citep{farrell2023}. In these interactions, AI tools can help states monitor and enforce their punitive economic regulatory policies. Finally, governments may also use AI to enhance their discursive power to shape economic policy preferences. For instance, governments may advance neoliberal logics of privatization through contracting more service provision to AI, such as welfare allocation, and normalizing these practices for the public. Governments could also use the AI race to encourage the public to support domestic firms.

Governments also decide the scope of rights and determine how these rights will be protected and enforced. Governments can expand instrumental power in the rights domain through using AI in law enforcement. Both judicial and policing agencies already make use of AI to augment and automate tasks such as evidence collection, evidence analysis, and judicial recommendations \citep{angwin2016, brayne2020, cordelli2020}. In terms of consolidating structural power against the governed, judicial and law enforcement agencies may also integrate AI tools within their organizational decision-making apparatus in ways that produce asymmetries for rightsholders, including embedding systemic practices like judicial case assignment and benchmarking judicial decisions against decisions made by AI \citep{taylor2023}. Finally, in discursive power, AI may be used  by governments to shape rights consciousness, that is citizens’ perceptions of the rights and processes to which they are entitled to by law. For instance, the use of AI in law enforcement might require real-time facial recognition tools that violate privacy, but governments may minimize how important privacy should be as a right to citizens inhabiting AI infused societies or use AI metrics delivering enhanced “public safety” as an adequate tradeoff. In this way, the rights and processes of the  governed may be further obfuscated.

\subsection{Private Authorities and Governance by AI}
While governments typically hold a monopoly on the legitimate use of violence, private actors such as organized criminal groups and security contractors also engage in violence. In traditional warfare, private military and security contractors may deploy autonomous weapons or AI surveillance tools with less accountability than there would be for government use \citep{srivastava2022b}. Mexican drug cartels have used malware against journalists while the militia Hezbollah and terrorist group Boko Haram have used cyber strikes against states \citep{handler2022}. Corporations also used AI tools to exert unequal structural power vis-à-vis the general public through the creation of for-profit predictive policing \citep{brayne2020} and spyware \citep{deibert2020} that are used jointly for commercial and governmental purposes, as seen with Palantir and NSO Group. Finally, AI can also be used to shape consumers’ threat perceptions of violence through making more “analytics” available without context, such as Amazon Ring video doorbell’s facial recognition system that identifies “strangers” or platforms like NextDoor and Facebook that amplify anecdotes about crime and violent extremism. In these examples, corporate goods and services that involve AI tools shape discourses of the prevalence and scope of societal violence, which may nudge public support for stronger law enforcement or counterterrorism practices.

Corporations exert instrumental power most obviously to shape the investment ecosystem by developing and deploying AI tools for profit or serving as consultants to help other firms navigate the AI “gold rush.” For instance, the International Chamber of Commerce has launched an AI system to advise its members, who are typically lobbyists, on negotiating trade agreements and contracts \citep[p.963]{arsenault2024}. Firms are also already using AI to increase their structural power over competitors through integration of AI tools throughout their business model \citep{wef2024}, as seen for instance in the adoption of algorithmic trading by global finance. Businesses also use AI to reshape supply chain management, for instance optimizing warehouse efficiency and shipping routes, to claim a better structural position. Moreover, corporate investment in AI helps maintain structural market dominance as the best performing AI models require large amounts of data, tremendous computing power, and elite human talent, all of which favor Big Tech firms who can then charge monopoly rents. In terms of discursive power, corporations may encourage consumers to view AI integration as an accepted practice of business and consumption. For instance, Ring may make it seem normal to consistently surveil one’s neighborhood to identify “potential threats.” The tremendous public response to ChatGPT as a new way of seeking information has led to a race to develop other chatbots and integrate them in existing services, such as Bing and Google Search, even though these models are not yet fully tested for accuracy and bias.
	
 Corporations also play a large role in how individual rights are respected. For instance, firms may violate employees’ privacy when AI tools are used for worker surveillance for instrumental purposes. Even innocuous-sounding wellness programs or social media apps lead to a “quantified” self \citep{ajunwa2023}. AI also enables corporations to exert structural power through automation in labor markets. Automation may impact how firms hire, for instance using AI tools in hiring processes. Under Title VII of the Civil Rights Act of 1964, employers cannot discriminate against applicants on the basis of sex, race, religion, disability, pregnancy, national origin, age, or genetic information. But predictive algorithms used in hiring processes have been found to systematically rank applicants lower on the dimensions of sex and race \citep{chen2023}. Integration of AI tools into organizational processes may then structurally advantage capital vis-a-vis labor as applicants lack knowledge about algorithmic bias and discrimination. Moreover, automation impacts whether firms hire at all by replacing humans with machines in software form such as tax preparation tools or hardware such as physical robots. Finally, AI advancements require data and corporations wield discursive power to alter the general perception of data rights in surveillance capitalism, where the proliferation of data-hungry platforms like Facebook, Google, Amazon, and Tencent seems like an inevitable trade-off \citep{zuboff2019}.

\subsection{Governance by Autonomous AI Agents}

Our discussion of AI’s uses for global governance has so far assumed states and corporations as the primary players. However, autonomous AI agents may also be seen as decision-makers themselves. Computer scientists are increasing the ability of AI systems to act like agents that pursue goals in diverse environments \citep{cheng2024, du2024}. The notion of AI as agents is related to other kinds of nonhuman agentic capacity. Both governments and corporations pursue goals, acquire resources, and perform actions in their environment even when these goals do not perfectly overlap with the humans that constitute them \citep{young2021}. In this way, governments and corporations are said to be agentic. Machine learning (ML) also creates agents that pursue goals, acquire resources, and perform actions in their environment. Many ML systems now perform tasks and play certain games at human capability and in many cases surpassing even the very best humans. In global governance, agency has been attributed to international organizations \citep{barnett2004}, non-governmental organizations \citep{keck1998}, corporations \citep{strange1996, hofferberth2019}, and regime complexes \citep{alter2018} in addition to states. Thus, the claim of agentic AI should be viewed along the same lines as other nonhuman agents in global governance such as bureaucracies, market logics, and international organizations. 

AI autonomy is generally viewed on a continuum of  humans-in-the-loop, humans-on-the-loop, and humans-out-of-the-loop, with each type broadly corresponding to humans initiating and being in control of a task, humans largely delegating to AI systems but retaining control over stopping a task, and humans taking no active role in a task. Table 2 summarizes how autonomous AI agents in humans-out-of-the-loop systems might be used to advance instrumental, structural, and discursive power relative to humans.

\begin{figure}[h!]
\centering
\includegraphics[width=\textwidth]{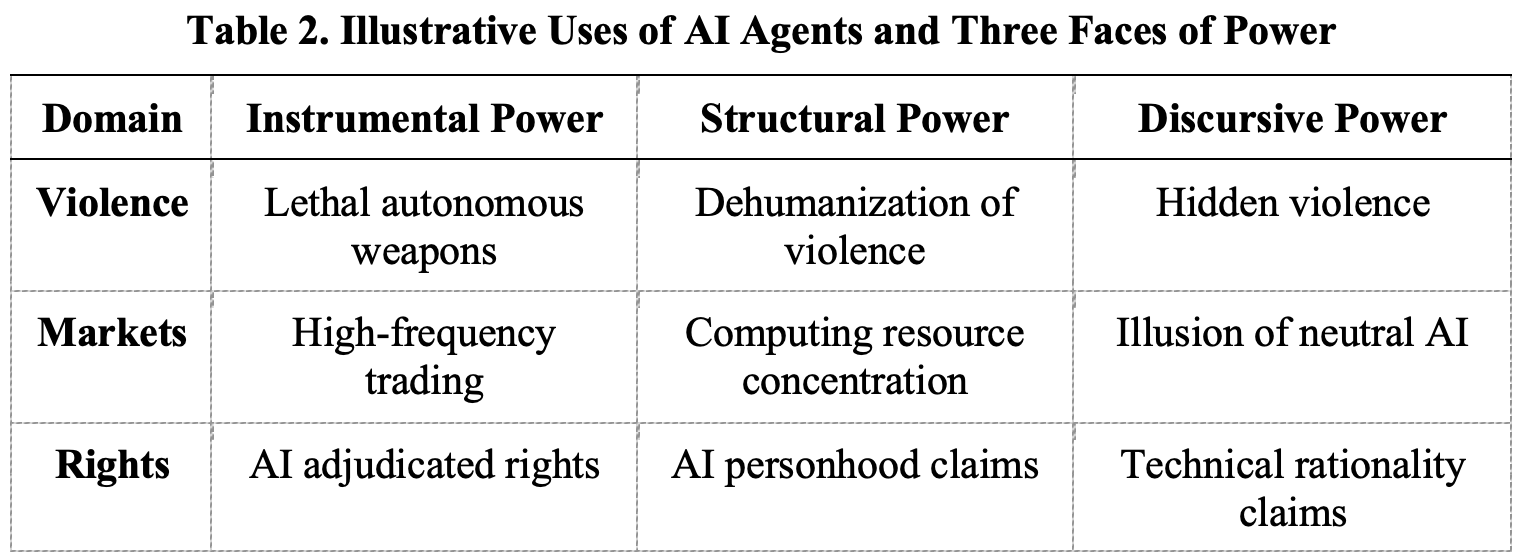} 
\end{figure}

Regarding instrumental power, lethal autonomous weapons (LAWS) likely already have the capability to identify a target and execute a violent attack \citep{etzioni2017, canfil2024}. LAWS have been “progressing” in their capacity to achieve the highest level of autonomy which involves no human action and the AI itself identifying targets and actively committing the violent act. AI tools may also be used to make automated strategic decisions on how to conduct violence with human decision-makers removed from the decision-making process. In structural power terms, the substitution of human planning for AI planning would systematically devalue human judgment and thus cede discretionary power to AI agents. Traditionally, some form of human deliberation is required to justify the use of violence, but this is less applicable to the deployment of violence by autonomous AI agents, which may lead to dehumanization of violence. Discursively, this likely shifts social expectations of how responsible humans feel about the violence, especially if autonomous AI systems provide human decision-makers with a sense of being further removed. If society is not exposed to the consequences of violence perpetrated in their name upon others, then the true cost of violence is hidden, which may desensitize publics to higher levels of violence or allow plausible deniability of particularly unjustifiable types of violence \citep{young2021}. 

Current autonomous AI agents cannot effectively execute all or even most market-relevant tasks, but within the domains that AI can execute these tasks they can be completed quickly and at scale for which human actors cannot compete \citep{korinek2022}. In instrumental terms, algorithmically executed high-frequency trading already shapes financial markets where “decisions to place bids to buy and offers to sell are made by algorithms, not humans” \citep[p.4]{mackenzie2021}. High-frequency trading is believed to have played a prominent role in the 2010 Flash Crash, which briefly led to a drop of 1,000 points on the Dow Jones Industrial Average. For autonomous AI agents accumulating structural power, we have already mentioned that access to computation is unevenly distributed. The computer chips required to train frontier AIs are currently maintained by a handful of large multinational corporations \citep{sastry2024}. Additionally, it is the AI models themselves that make the most abundant and efficient use of these computational resources, giving them a comparative advantage vis-à-vis humans. The increasing capabilities of AI agents also presents the opportunity for excessive automation, which would provide further structurally advantage AI agency in making decisions related to humans. Growing use of autonomous AI agents affords discursive power for organizing markets as well. For instance, the rise of global capital has reshaped cultural perceptions of what is appropriate in market relations. The cultural logic of capital places value on profit, competition, productivity, and efficiency, under the illusion of the “invisible hand,” and the cultural logic of bureaucratization privileges roles, responsibilities, hierarchy, performance, and efficiency, under the illusion of “rational procedures” \citep{bullock2022}. Similarly, the cultural logic of automation would usher its own set of market values related to speed, efficiency, fairness, and labor redundancy, all the while cloaked under the illusion of “neutral AI.”

Finally, the use of autonomous AI agents can alter rights in instrumental ways. In full automation, the rights of displaced workers would have to be adjudicated. But even in partial automation, rights are directly implicated, such as the due process rights of wrongly-identified individuals as a result of autonomous facial recognition programs deployed by law enforcement \citep{press2023}. AI agents may make autonomous judicial decisions, “resolving” backlogs, such as in asylum cases. In structural power terms, rights are impacted through both weakening the bargaining power of labor and treating autonomous AI models on par with other ethical agents. Just as corporations and institutions have been afforded ethical relevance, autonomous AI systems may be viewed as “ethical impact agents” \citep{livingston2019} or would-be “synthetic moral agents” \citep{erskine2024}. Moreover, just as corporate personhood claims can distort money into speech in American campaign finance, any future claims to AI legal personhood also opens pathways of structural imbalance \citep{bryson2017}. Autonomous AI agency will further shape social discourse on the balance of rights across AI agents and human agents. On the one hand, there are claims that “technical rationality” would result in less bias and discrimination as machine inferences are not swayed by human-centric biases and emotions. In this context, we could imagine some AI systems as rights protecting. On the other hand, autonomous AI agents may still operate from particular assumptions that are rights diminishing. As rights clash, for instance privacy and safety, there are concerns that AI agents’ assumptions may be at odds with human values or that AI agents may not understand the desire for particular rights, such as right to dignity, without experiencing the corresponding struggles when those rights are deprived. 

We have so far mapped how public and private global governors may use AI systems to organize violence, markets, and rights through instrumental, structural, and discursive power. We also considered global governance by autonomous AI agents. We next move to analyzing the implications of AI’s multifaceted global governance roles for state sovereignty.

\section{Implications for Digital Sovereignty}
This section suggests a few points of analysis for how the use of AI technologies by global governors implicates sovereignty. Sovereignty is a contested concept in IR theory, at once indispensable yet subject to persistent debate about its meaning  \citep[p.607]{wendt2008},\citep[p.455-456]{paris2020},\citep[p.12-15]{srivastava2022b}. Rather than offer one definition of sovereignty, we center our discussion around two prevailing interpretations: sovereignty as an institution and sovereignty as a practice. 

\subsection{Sovereignty as Institution}
What separates states from other entities is the shared international understanding that only states represent legitimate political authority. That legitimacy, in turn, relies on claims about sovereign authority that are distinct from claims about other kinds of authority, such as religious authority. In traditional understanding, sovereignty is “the idea that there is a final and absolute political authority in the political community … and no final and absolute authority existed elsewhere” \citep[p.365]{hinsley1986}. Here, sovereignty is an idealized claim – i.e., “the idea that” – not an accurate representation of reality. It is in this context that scholars observe, “state control has waxed and waned enormously over time, regions, and issue-areas while the state's claim to ultimate political authority has persisted for more than three centuries” \citep[p.214]{thomson1994}. The purpose of making idealized claims about sovereignty is “to express and realize the principles that make a state a particular state” \citep[p.13]{inayatullah1995}. Variously referred to as international legal sovereignty \citep{krasner1999} or de jure sovereignty, in this interpretation “when states recognize each other’s sovereignty as a right then we can speak of sovereignty not only as a property of individual states, but as an institution shared by many states” \citep[p.280]{wendt1992}. In short, sovereignty as an institution constitutes states as states \citep[p.430-431]{onuf1991}. 

In a “sovereignty cartel” \citep{barkin2021}, states jealously guard and uphold the international institution of sovereignty to keep out nonstate actors or quasi states, no matter how powerful. Big Tech firms, like oil companies before them, are wealthier than most states, but none can be a member of the United Nations or engage in sovereign lending or sign peace treaties. But the sovereignty cartel faces challenges. Early globalization debates centered partially on whether borderless corporations would result in disempowered states and the erosion of state sovereignty \citep{strange1996, goldsmith2006}. Similarly, scholars speak of “sovereignty costs” to indicate the tradeoffs of accepting another’s authority in place of or alongside one's own \citep{krasner1999}. For instance, in the European Union, member states give up some of their individual authority claims to make monetary policy or control borders when pooling sovereign authority. As such, joining the EU may come with some sovereignty costs. But as a supranational organization, the EU is made up of other sovereign states. In that regard, the international institution of sovereignty is not necessarily degraded. Moreover, to the extent the EU boosts member state capacity that would not be possible absent joining the organization, it might end up being a “sovereignty boon.” As such, there are no straightforward ways to assess how sovereignty as an institution may be implicated by changes in global power dynamics. In the context of AI, state actors may use AI to further entrench their sovereign claims. There are two ways this may happen. 

First, states may use the threat of powerful private interests to assert more sovereign control over digital infrastructures vis-a-vis any corporate sovereign challengers \citep{bellanova2022}. Modern frontier AI systems are expensive endeavors. To collect massive training data, procure superclusters of computation, and have the human talent required to develop the requisite training and inferential algorithms requires large amounts of capital. Additionally, until the past few years, it was thought that these types of AI systems would be unable to outperform symbolic, knowledge-based models. Consequently, well-resourced private actors have created the new wave of generative AI systems. Thus, the current frontier models are created by private companies and are designed to be useful agents in the marketplace for which the companies can deploy at profit. This suggests that these models will be designed to compete with labor and seek capital returns. As a result, the most powerful cutting-edge AI systems are, generally speaking, private artifacts. Within this context, Nvidia’s CEO told governments that they must invest in “sovereign AI” as “it codifies your culture, your society’s intelligence, your common sense, your history–you own your own data” \citep{nvidia2024}. Separately, European policymakers have presented American tech firms as a threat to the EU’s “digital sovereignty” \citep{broeders2023}. Europe has passed the strongest regulations related to AI so far. Even the usually pro-innovation American lawmakers have held meetings with AI firms to secure voluntary commitments related to public safety. China too has moved to regulate its tech firms, for instance by forcing the ouster of Jack Ma, the former CEO of Alibaba, and restricting training data for generative AI models.

Second, in “AI nationalism” states may use interstate rivalry in an AI race to exert more sovereign overreach. While the current wave of technological innovation is predominantly found in the private sector, state interest to claim AI for itself is significant. Even though the exact amounts of government investments in foundation models in the U.S. and China are unknown, from public discourse it is apparent that both states view AI as central to their competitive advantage. China has declared it will be an AI superpower by 2030, aiming for a one trillion renminbi industry. In the U.S., Congress has taken note of AI’s role in the Sino-American rivalry and moved to secure domestic supply chains related to supercomputing. The CHIPS Act, passed in 2022, makes it difficult for Chinese firms to procure powerful chips that underpin high-performing AI models. A current bill in Congress would extend the restrictions to exporting AI software, not just hardware. The U.S. has incentivized its European and Middle Eastern allies to also block China \citep{farrell2023}. Yet, Chinese AI advances are continuing, in particular because of an open-source approach that allows developers to build on top of each other much more quickly than in closed systems such as OpenAI’s \citep{tobin2024}. China’s version of “digital sovereignty,” then, is a response to both American public and private power in AI. While Europe aims to regulate private companies, European digital sovereignty also embraces geoeconomics, an ideology that views championing European firms as central to digital progress. The French have bemoaned: “We must rise to the challenge of A.I., or risk losing the control of our future” \citep[p.3]{aicommission2024}. The U.S., for its part, has moved to articulate “digital solidarity” as a counter to the Chinese and European sovereignty narratives. 

The presence of challengers in the AI space provides states with reasons to defend their idealized claims to monopolizing legitimate political authority. In the AI context, this has meant some articulations of “digital sovereignty” that reassert a state (or in European case, suprastate) claim to sovereign authority in response to challengers. It is an open question whether states will succeed in reality, but AI proliferation in global governance does not signal the end of sovereignty as an international institution.

\subsection{Sovereignty as Practice}
Sovereignty as an institution might represent a state-centric claim to supreme political authority, but sovereignty is messier in practice. States may not be functionally in charge of making sovereign decisions, thus lacking de facto (in fact) sovereignty. The functions we assign to sovereign powers may also evolve over time. While we traditionally expect sovereigns to collect taxes, fight wars, write laws, and defend rights, among other activities, these roles evolve. For instance, trash collection was not a sovereign function in the early or late modern era, but is widely seen as one today. Taking care of wounded soldiers went from nonexistent before the 19th century to an integral part of international humanitarian law in the 20th century. In this interpretation of sovereignty, “the sovereign state is an ongoing accomplishment of practice, not a once-and-for-all creation of norms that somehow exist apart from practice” \citep[p.413]{wendt1992}. As such, the “‘traditional’ meaning of sovereignty is not as foundational and timeless as is commonly assumed” \citep[p.79]{glanville2013}, \citep{ashley1984}. In approaching sovereignty as practice, scholars “are interested in considering the variety of ways in which states are constantly negotiating their sovereignty” \citep[p.11]{biersteker1996}.
	
 Research that regards sovereignty as practice shows that a variety of nonstate entities carry out functions associated with sovereign states \citep{lake2003, slaughter2005, doty2007, cooley2009, avant2010, best2014, phillips2015, srivastava2022b}. Contractors fight wars, lobbyists conduct commerce, and nongovernmental organizations deliver welfare. While nonstate expressions of sovereign power may invite discussions of sovereignty costs or public versus private, some view sovereign competence as a joint enterprise between public and private entities. For instance, a “states and markets” perspective pushes back against the early globalization debates on “states versus markets” to argue that powerful corporations can enhance powerful states \citep{mikler2018}. Relatedly, in a framework of “hybrid sovereignty,” states do not become “less” sovereign by collaborating with private actors; instead, sovereign power itself changes to include both public and private sources \citep{srivastava2022a}. 

AI’s many roles in global governance implicates digital sovereignty as practice in two ways. First, public and private global governors may jointly pursue AI advances that affect their capacities. Governments who seek to be empowered by AI also need to rely on public/private collaborations. Consider that while one plank of EU’s digital sovereignty approach targets American tech companies for regulation, another plank partners with them for establishing its “sovereign cloud” capabilities \citep{broeders2023}. As such, European digital sovereignty is akin to “performative discourse” \citep{adler2024}, meant to include a diverse range of policy practices. As European data protection laws dictate how digital data is treated around the world, the EU also passed new regulations that make it easier to access public data for commercial purposes. While government programs such as PRISM reveal the shadow links between American firms and intelligence agencies for mass surveillance, many public-private hybrid collaborations occur in plain sight. Smart cities rely on public-private data infrastructures to direct traffic, pick up trash, or fix potholes. Many airports now feature facial recognition boarding tools. Contact tracing by health ministries during the early part of the Covid pandemic relied on Apple and Google’s operating systems. Amazon Ring cameras feed into law enforcement systems for neighborhood surveillance. In this way, public attempts at digital sovereignty may not be practiced against private interests, but decidedly with them. 

Second, AI systems may come to represent new ways of practicing sovereignty. We have already discussed the many ways in which AI systems may empower public and private governors organizing violence, markets, and rights. Through critical functional embeddedness, decision-making logics of AI systems could permeate how governors make decisions as well. For example, greater use of AI technologies in governance may create new incentives for speed and scale or reduce demands for privacy protection. In embracing “sovereign AI,” the Indian government recognizes: 

We can take two options. One is to say, as long as there is an AI ecosystem in India whether that is driven by Google, Meta, Indian startups, and Indian companies, we should be happy about it. But we certainly don’t think that is enough. With sovereign AI and an AI compute infrastructure … the government is not looking to just compete with the generative AI type of model. It also wants to focus on real-life use cases in healthcare, agriculture, governance, language translation, etc., to maximise economic development \citep{barik2023}. 

The first option recalls sovereignty as an institution where the goal is for a self-sufficient AI ecosystem or at least one where the state has control over the AI ecosystem. The second option reflects sovereignty as practice where AI infrastructures transform how sovereign governance happens in areas such as healthcare, agriculture, and language translation. The point about language is especially relevant for global governance as instantaneous AI translation might enable more integrated real-time communications for policymaking, including among rivals \citep[p.890]{ding2024}.  

In addition to influencing how governors make decisions, AI agents may make decisions that do not represent the interests of public or private governors. One risk identified by the AI Safety field is the alignment problem where an AI agent's goals may not be aligned with that of its designers and deployers \citep{christian2021, ngo2022}. If the alignment problem is not solved, then AI agents may pursue goals that are different from the companies or states that create them. Another risk is that AI agents outcompete human agents on general tasks, a possibility amplified by the evolution of generative AI \citep{lazar2024}. This is related to the control problem in AI Safety, where if AI agents can outcompete human agents on general sets of tasks, then the human agents may not be able to meaningfully control the behavior of advanced AI agents \citep{russell2019, bengio2024}. The alignment and control problems present the challenge that AI agents may pursue goals that neither public nor private governors control. As such, “sovereign AI” takes an entirely different connotation than the one we have discussed so far. Unaligned or uncontrolled AI agents could make cooperation between states, corporations, and AI agents exceedingly difficult \citep{bullock2024, drexler2019, hendrycks2023}. 

Ultimately, the dynamics of AI’s use in global governance generates new ways of practicing digital sovereignty, including further incentivizing public/private collaboration and opening possibilities for AI agency, including AI agents as new forms of global governors. However, one lesson from global governance is that while public/private hybrids may produce or prop up sovereign power in practice, they are less successful at projecting sovereign authority or gaining entry into the sovereignty cartel \citep{srivastava2022b}. If hybrid practices and governance by AI become fully embedded, then we should expect to see states attempt to reconcile with ideas expressed in sovereignty as an institution, such as by invoking more claims of state supreme authority over AI even though state power is maintained through AI.

\section{Conclusion}
This essay mapped AI’s multifaceted role in global governance. As a general purpose technology, AI systems are used by both public and private governors to enhance their instrumental, structural, and discursive power in the domains of violence, markets, and rights. We provided a taxonomy of AI power payoffs using examples of relevance to global governance broadly. We also assessed how digital sovereignty is implicated in AI-empowered global governance efforts by presenting different analyses of states seeking sovereign control over AI infrastructures and establishing sovereign competence through AI infrastructures. Rather than foreseeing a technopolar order where technology companies replace states, we argued that AI systems will embed in global governance to create dueling dynamics of public/private cooperation and contestation. These dynamics are not unique to AI, but AI offers a particularly fertile issue area for studying overlapping agency in global politics. 

Moving forward, one strand of IR research should explore what new cooperative and competitive strategies arise between public and private governors in sovereignty games as they acquire power and aim to secure or hold onto authority. This research could focus on whether and how private actors are accommodated, if at all, in AI regulation. Moreover, detailed studies on AI private governance (such as standard-setting) or comparative AI firm strategies, for instance between American and Chinese technology sectors, would also be valuable. Broadly, we also need to know more about how intergovernmental organizations such as the UN or civil society groups respond to AI advances in the private sector. 

Another set of research inquiries concern geopolitical dynamics between states, where questions include how AI exacerbates or ameliorates inequalities between the global North and South. AI is sometimes considered a democratizing technology, where, for instance, more people have access to the powerful technologies of ChatGPT and the like than ever before. However, frontier AI development is still concentrated in a handful of states and companies. Moreover, many states have signaled interest in developing safe and trustworthy AI, for instance through endorsing OECD and UNESCO guidelines on ethical AI. But under what conditions do states regulate AI in rights-protecting ways? Relatedly, will state regulation of AI move towards cohesion or fragmentation? 

Finally, IR scholars should also pay attention to changing AI capabilities. We have discussed the possibility of AI agents exerting significant power over governance decision-making. Researchers should explore the ethical and political implications of AI agency, including the potential for autonomous AI to alter the power dynamics between states and corporations and even the meaning of governance altogether.

\newpage
\bibliographystyle{apalike}  
\bibliography{references2}

\end{document}